\documentclass[letterpaper]{article} 
\usepackage{aaai24}  
\usepackage{times}  
\usepackage{helvet}  
\usepackage{courier}  
\usepackage[hyphens]{url}  
\usepackage{graphicx} 
\usepackage{natbib}  
\usepackage{caption} 
\usepackage{algorithm}
\usepackage{microtype}
\usepackage{subfigure}
\usepackage{booktabs} 
\usepackage{amsmath}
\usepackage{amssymb}
\usepackage{mathtools}
\usepackage{amsthm}
\usepackage{enumerate}
\usepackage{nicefrac}
\usepackage{amsfonts}
\usepackage{dirtytalk}
\usepackage{multirow}
\usepackage{enumitem}
\usepackage[title]{appendix}
\usepackage{algpseudocode}
\usepackage{xcolor}
\usepackage{placeins}

\usepackage[capitalize]{cleveref}
\crefname{section}{Sec.}{Secs.}
\Crefname{section}{Section}{Sections}
\Crefname{table}{Table}{Tables}
\crefname{table}{Tab.}{Tabs.}
\usepackage{newfloat}
\usepackage{listings}
\DeclareCaptionStyle{ruled}{labelfont=normalfont,labelsep=colon,strut=off} 
\floatstyle{ruled}
\newfloat{listing}{tb}{lst}{}
\floatname{listing}{Listing}
\title{Contrastive Credibility Propagation for Reliable Semi-supervised Learning}
\author {
    Brody Kutt,
    Pralay Ramteke,
    Xavier Mignot,
    Pamela Toman,
    Nandini Ramanan,
    Sujit Rokka Chhetri,
    Shan Huang,
    Min Du,
    William Hewlett
}
\affiliations {
    Palo Alto Networks, AI Research\\
    \textrm{\{bkutt, pramteke, xmignot, ptoman, nramanan, schhetri, shuang1, midu, whewlett\}@paloaltonetworks.com}
}

\begin{document}

\maketitle

\begin{abstract}
Producing labels for unlabeled data is error-prone, making semi-supervised learning (SSL) troublesome. Often, little is known about when and why an algorithm fails to outperform a supervised baseline. Using benchmark datasets, we craft five common real-world SSL data scenarios: few-label, open-set, noisy-label, and class distribution imbalance/misalignment in the labeled and unlabeled sets. We propose a novel algorithm called Contrastive Credibility Propagation (CCP) for deep SSL via iterative transductive pseudo-label refinement. CCP unifies semi-supervised learning and noisy label learning for the goal of reliably outperforming a supervised baseline in any data scenario. Compared to prior methods which focus on a subset of scenarios, CCP uniquely outperforms the supervised baseline in all scenarios, supporting practitioners when the qualities of labeled or unlabeled data are unknown.
\end{abstract}

\section{Introduction}
\label{sec:intro}

A fundamental goal of semi-supervised learning (SSL) is to ensure the use of unlabeled data results in a classifier that outperforms a baseline trained only on labeled data (supervised baseline). However, this is often not the case \cite{NEURIPS2018_c1fea270}. The problem is often overlooked as SSL algorithms are frequently evaluated only on clean and balanced datasets where the sole experimental variable is the number of given labels. Worse, in the pursuit of maximizing label efficiency, many modern SSL algorithms such as \cite{DBLP:journals/corr/abs-1911-09785,https://doi.org/10.48550/arxiv.2001.07685,simmatch,Li2021CoMatchSL} and others rely on a mechanism that directly encourages the marginal distribution of label predictions to be close to the marginal distribution of ground truth labels (known as distribution alignment). This assumption is rarely true in practice. We identify five key dataset quality variables (data variables) that can strongly impact SSL algorithm performance and are common in real-world datasets.

\begin{enumerate}
    \item \textbf{Few-label}: Varying the number of labeled samples per class typically while the amount of unlabeled data grows or is held constant.
    \item \textbf{Open-set}: Including and varying the ratio of out-of-distribution (OOD) samples, \textit{i.e.} samples which belong to no class, in the unlabeled data.
    \item \textbf{Noisy-label}: Varying the percent of given labels that are incorrect.
    \item \textbf{Class distribution imbalance/misalignment}: Varying the disparity of class frequency distributions between labeled and unlabeled data. We explore increasingly imbalanced distributions in the labeled and unlabeled data separately (while keeping the other uniform to ensure misalignment).
\end{enumerate}

Thoroughly evaluating the practicality of an SSL algorithm requires analyzing its response to these variables at differing severity. Identifying specific quality issues in a dataset can be challenging for practitioners. Often, real-world SSL workflows encounter scarcity and noise in labeled data, which is also sampled from a distribution distinct from the unlabeled data, derived from the often unknown target distribution. Moreover, the unlabeled data distribution may be inaccessible due to privacy concerns or real-time data collection in fully autonomous systems. Consequently, such systems necessitate external components like Out-of-Distribution (OOD) detectors to prevent failures, albeit at the cost of increased complexity. Instead of maximizing the robustness to any one data variable, \textit{we strive to build an SSL algorithm that is robust to all data variables,} \textit{i.e.} \textit{can match or outperform a supervised baseline.} To address this challenge, we first hypothesize that sensitivity to pseudo-label errors is the root cause of all failures. This rationale is based on the simple fact that a hypothetical SSL algorithm consisting of a pseudo-labeler with a rejection option and means to build a classifier could always match or outperform its supervised baseline if the pseudo-labeler made no mistakes. Such a pseudo-labeler is unrealistic, of course. Instead, we build into our solution means to work \textit{around} those inevitable errors.

Our contributions can be summarized as follows. To the best of our knowledge, our work is the first to 1) define an SSL algorithm that unifies a pseudo-labeling strategy and an approach to overcome label noise 2) use credibility vectors to properly represent uncertainty during pseudo-label generation 3) propose a generalized contrastive loss for non-discrete positive pairs 4) demonstrate a reliable performance boost over a supervised baseline across five real-world data scenarios at differing levels of severity. The rest of this work is organized as follows. We overview related works in \cref{sec:related_work}. We introduce CCP in \cref{sec:ccp}. In \cref{sec:experimental_results}, we detail our experimental results before concluding in \cref{sec:conclusion}.

\section{Related Work}
\label{sec:related_work}

SSL has a rich history in AI research \cite{DBLP:journals/corr/abs-2103-00550,books/mit/06/CSZ2006}. We focus on two dominant approaches. These are pseudo-labeling and consistency training. CCP draws inspiration from both approaches.

\subsubsection{Pseudo-labeling}
\label{subsubsec:pseudo_labeling}

These methods typically constitute transductive learning \cite{shi2018transductive,iscen2019label}. Here, the objective is to generate proxy labels for unlabeled instances to enhance the learning of an inductive model. Seminal work in \cite{lee2013pseudo} simply uses the classifier in training to produce pseudo-labels inductively which are trained upon directly. This is problematic in that it actively promotes confirmation bias \textit{i.e.} the model will learn to confirm its predictions. This was later extended to include a measure of confidence in \cite{shi2018transductive}. Closely related is the concept of self-training which iteratively integrates into training the most confident of these pseudo-labeled samples and repeats \cite{dong-schafer-2011-ensemble,https://doi.org/10.48550/arxiv.2109.00778,9156610}. These techniques can become unstable when pseudo-label error accumulates across iterations. More recent work has addressed accumulating errors by using independent models to utilize pseudo-labels \cite{https://doi.org/10.48550/arxiv.2202.07136}. LPA \cite{lpa} is a popular graph-based technique for generating pseudo-labels transductively but has many failure cases \cite{cred_assessment}. Much work tends to use LPA as a transductive inference mechanism for pseudo-labeling. Such inferred pseudo-labels are highly noisy and thus problematic for SSL. The CCLP algorithm \cite{pmlr-v80-kamnitsas18a} tries to circumvent this by instead using LPA-derived pseudo-labels only for graph-based regularization of the encoder. Like CCP, \cite{wang2022r2} proposed repeatedly re-predicting pseudo-labels during an optimization framework akin to self-training. Despite the popularity of pseudo-labeling-based SSL approaches, they tend to break down when faced with a large amount of pseudo-label errors.

\subsubsection{Consistency Training}
\label{subsubsec:consistency_training}

Analogous to perturbation-based SSL and contrastive learning, these techniques train a network to produce a single, distinct output for a sample under different augmentations (transformations). Work in \cite{xie2020unsupervised} minimizes a consistency loss for unlabeled data and a standard classification loss for labeled data simultaneously which could introduce irreconcilable gradients. Other work \cite{pmlr-v119-chen20j,https://doi.org/10.48550/arxiv.2006.10029}, shown superior to \cite{xie2020unsupervised} in SSL, employ an unsupervised consistency-like loss in a pretraining stage before training solely with given labels. Self-supervised consistency loss is attractive in that it eliminates supervision errors but lacks power in that class structure from unlabeled data is never leveraged directly. Instead, CCP introduces a \textit{softly} supervised consistency (contrastive) loss with true labels and pseudo-labels.

\section{Contrastive Credibility Propagation (CCP)}
\label{sec:ccp}

Here we present our method for transductive pseudo-label refinement and classifier training called Contrastive Credibility Propagation (CCP). We establish notation, provide a high-level overview, introduce credibility vectors, formalize our loss functions, detail the CCP iteration, introduce our subsampling procedure, and then explain how to build a classifier on CCP-derived pseudo-labels. 

\subsection{Notation}
\label{subsec:notation}
Consider a classification task among a set of $K$ classes $c=[0, 1, \ldots, K - 1]$. Consider a partially labeled dataset $X$. Denote all indices to labeled (unlabeled) samples as $L$ ($U$). Denote $I=L \cup U$. Each labeled sample $l \in L$ has a known one-hot class vector $y_l \in [0, 1]^{K}$ with a 1 (0) in the on (off) position. We maintain a real-valued credibility vector $q_i$ of length $K$ for each sample $i$: $q_i \in [-1, 1]^{K}$ (see \cref{subsec:credibility}). In training, denote the indices of a random batch of size $n$ as $\mathcal{B} \subseteq I$. We decompose $\mathcal{B}$ into the indices of labeled, $\mathcal{B}_l$, and unlabeled, $\mathcal{B}_u$, samples. We define a set of data transformations, $\mathcal{T}$. We draw two transformations, $t_1, t_2 \in \mathcal{T}$, randomly for every batch. A transformed pair will share one $q_i$. A batch of transformed data and credibility vector tuples will be of size $2n$ denoted $\{(x_i, q_i), (x_{i+n}, q_i)\}_{i\in \mathcal{B}}$ where $i$ and $i+n$ denote the indices of a transformed pair. 

\subsection{Overview}
\label{subsec:overview}

\begin{figure}[tb]
  \centering
  \includegraphics{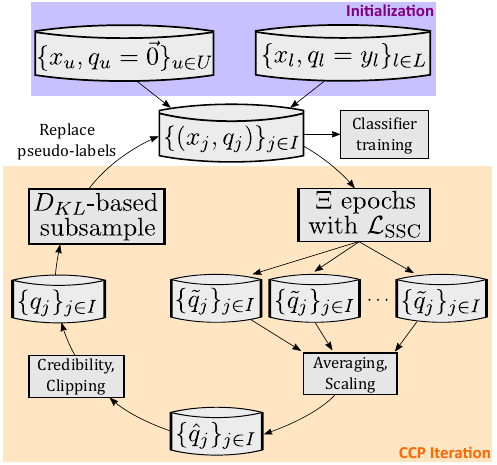}
  \caption{The credibility vectors for all samples, $\{q_j\}_{j\in I}$, are initialized with the given labels, label guesses, or $\vec{0}$ if unknown. For trusted labeled data, the associated $q_l$ may be permanently clamped to that value. A single CCP iteration first transductively generates new $\{\tilde{q}_j\}_{j\in I}$ for each of $\Xi$ epochs of training via \cref{q_j} while minimizing $\mathcal{L}_{\text{SSC}}$. All $\{\tilde{q}_j\}_{j\in I}$ are averaged and scaled such that the maximum of the strongest credibility vector is $1$ to form $\{\hat{q}_j\}_{j\in I}$. We then perform a final credibility adjustment and clip all values outside $[0, 1]$ to form $\{q_j\}_{j\in I}$. Once pseudo-labels have been refined through CCP iterations, they are then used to build an inductive classifier while minimizing both $\mathcal{L}_{\text{SSC}}$ and $\mathcal{L}_{\text{CLS}}$. The (inherently error-prone) results of \cref{q_j} are never used to directly supervise error-sensitive contrastive and classifier losses.} 
  \label{ccp_overview}
\end{figure}

CCP aims to equip an SSL training process to be robust to errors in the pseudo-labels assigned to unlabeled data. CCP thus consists of iterative pseudo-label refinement \textit{before} building a classifier. When combined with our credibility representation, this refinement helps to nudge pseudo-labels in the direction of the true class and nullify the data for which true labels are unknowable or ambiguous. To do this, we merge a process of transductively propagating pseudo-labels (in the form of credibility vectors) and an outer loop built for overcoming instance-dependent noisy labels based on the SEAL algorithm \cite{https://doi.org/10.48550/arxiv.2012.05458}. To train a transductive pseudo-label generator, we define a generalization of unsupervised (SimCLR \cite{pmlr-v119-chen20j}) and supervised (SupCon \cite{https://doi.org/10.48550/arxiv.2004.11362}) contrastive loss that can make use of non-discrete, \textit{i.e.} uncertain, positive pairs. We also define a pseudo-label subsampling strategy that limits the divergence of pseudo-label class distributions before and after subsampling which we hypothesize to be the primary cause of instability of similar subsampling procedures. We show it can lead to higher pseudo-label accuracy and faster convergence, especially with increasingly unclean data. Further details are provided in \cref{ccp_overview}.

\begin{figure}[tb]
  \centering
  \includegraphics{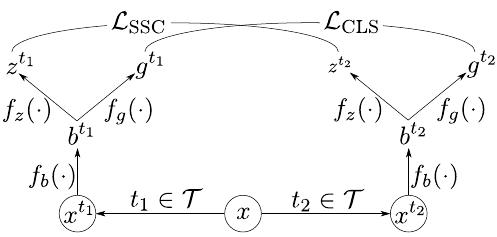}
  \caption{CCP uses an encoder $f_b(\cdot)$ and two projection heads: $f_z(f_b(\cdot))$ for contrastive loss and $f_g(f_b(\cdot))$ for classification loss.}
  \label{ccp_arch}
\end{figure}

The network architecture consists of an encoder $f_b$, that computes a vector encoding $f_b(x)=b$. One linear projection head $f_z$ computes an encoding used for contrastive learning, $f_z(b)=z$. A separate linear projection head, $f_g$, computes $f_g(b)=g\in \mathbb{R}^{K}$ for a classification loss. Attaching $f_g$ and $f_z$ to $f_b$ is motivated by \cite{pmlr-v119-chen20j,https://doi.org/10.48550/arxiv.2006.10029,https://doi.org/10.48550/arxiv.2004.11362}. These architectural components are illustrated in \cref{ccp_arch}. Between iterations and before classifier training, we reset the variables in $f_b$, $f_z$ back to a prior state (either a random initialization or pretrained using $\mathcal{L}_{\text{SSC}}$'s unsupervised counterpart, SimCLR).

Motivated by \cite{pmlr-v119-chen20j,https://doi.org/10.48550/arxiv.1803.11175}, the similarity function we use is angular similarity defined by $\phi(z_i, z_j) = 1-\nicefrac{\arccos\left(\frac{z_i \cdot z_j}{\lVert z_i \rVert \lVert z_j \rVert}\right)}{\pi}$.

\subsection{Credibility Adjustments}
\label{subsec:credibility}

\begin{figure}[tb]
  \centering
  \includegraphics[scale=0.5]{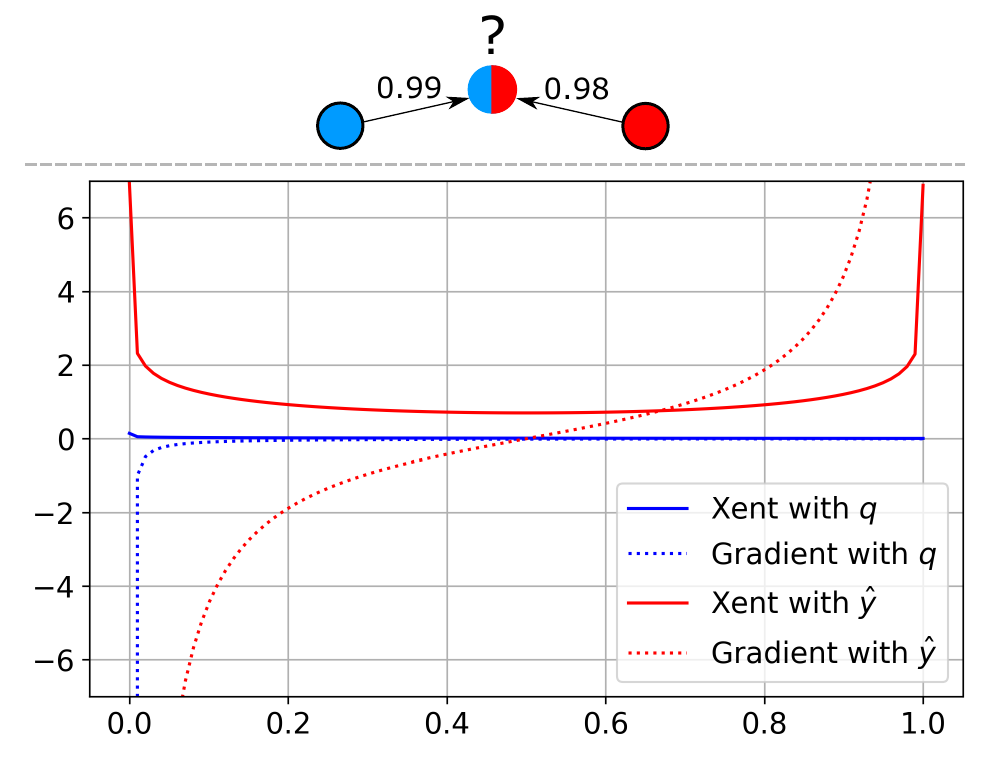}
  \caption{\textit{Top:} An unlabeled sample features high similarity scores of 0.99 and 0.98 for both classes. A conventional softmax label is computed as $\hat{y}=[0.502,0.498]$. A credibility vector, $q$, is computed as $0.99-0.98=0.01$ and $0.98-0.99=-0.01$ (clipped to 0 before it's used in Xent). Thus, $q=[0.01,0.0]$. \textit{Bottom:} Consider $x$ to be the softmax output of a binary classifier for the blue class. Using $\hat{y}$ to compute Xent induces strong gradients at either pole despite the true class being nearly ambiguous. Using $q$ will ensure the gradient for this sample is near $0$ everywhere except $x=0$. If $q=\vec{0}$, \text{i.e.} the sample is equally similar to both classes, the gradient is $0$ everywhere.}
  \label{cred_vecs}
\end{figure}

The CCP algorithm uses credibility vectors, $q_i$, in place of one-hot or softmax label vectors. Credibility vectors are more expressive. A credibility score of $1$ means maximum confidence in class membership, $0$ means uncertainty, and $-1$ means maximum confidence in class non-membership. \textit{Each credibility score given a sample and class is the similarity between that sample and class minus the highest similarity to all other classes}. Credibility adjustments are formalized in \cref{ccp}, \cref{ccp:begin_cred_adj,ccp:cred_adj,ccp:end_cred_adj} and \cref{q_j}, \cref{q_j:begin_cred_adj,q_j:cred_adj,q_j:end_cred_adj}. The core idea of credibility, similar to \cite{cred_assessment}, is to extend similarity measurements to capture ambiguity brought forth by competing similarities \textit{i.e.} conditional similarity. Unlike \cite{cred_assessment}, we represent credibility in our label vectors for use throughout the algorithm. For trusted labeled data, credibility vectors, $\{q_l\}_{l\in L}$, are clamped with 1 (-1) in the on (off) position. Before applying credibility vectors to $\mathcal{L}_{\text{SSC}}$, $\mathcal{L}_{\text{CLS}}$, and \cref{q_j}, we clip to a $[0,1]$ range. A clipped credibility vector thus consists of $0$ everywhere except the strongest value which is scaled down by the second strongest value. However, negative values are still useful when averaging across epochs in \cref{ccp:avg_over_epoch}. The single non-zero value reflects the strength (confidence) of the label prediction. At initialization, unlabeled data receive $q_u=\vec{0}$ (maximum uncertainty). Throughout CCP, the influence of $x_i$ scales with the magnitude of the non-zero value in $q_i$ ($q_i=\vec{0}$ will ensure $x_i$ exhibits no effect anywhere). Because CCP is designed to work with noisy (pseudo-)labels, if uncertain label guesses exist, these can be used for initialization instead. \cref{cred_vecs} provides a concrete example of a credibility vector calculation and its effect on a cross-entropy (Xent) gradient.

\subsection{Soft Loss Functions}
\label{subsec:soft_loss_functions}

$\mathcal{L}_{\text{SSC}}$ and $\mathcal{L}_{\text{CLS}}$ are contrastive and classification loss functions, respectively. Both are supervised by $\{q_j\}_{j\in I}$ provided by the CCP algorithm. Each sample in a transformed pair has identical credibility vectors and is treated independently. Thus, to simplify the notation, we formalize $\mathcal{L}_{\text{SSC}}$ and $\mathcal{L}_{\text{CLS}}$ for a batch of data and (clipped) credibility vector tuples, $\{(x_i, q_i)\}_{i\in \mathcal{B}}$.

We define an $n \times n$ pairwise matching matrix, $M$, where $m_{i,j} = q_i \cdot q_j$ for $i,j \in \mathcal{B}$. Each row of $M$ contains the weights for a weighted arithmetic mean of normalized contrastive losses for that sample to all other samples based on the evidence of a positive pair relationship. We scale pairwise similarities by temperature $\tau$ and take the exponential as in \cite{pmlr-v119-chen20j,https://doi.org/10.48550/arxiv.2006.10029,https://doi.org/10.48550/arxiv.2004.11362} to form an $n \times n$ matrix $A$ defined by $a_{i, j} = \exp(\phi(z_i, z_j)/\tau)$ for $i,j \in \mathcal{B}$. We construct a strength vector, $\omega$, defined by $\omega_i= \max(q_i)$ for $i \in \mathcal{B}$. Each $\omega_i$ serves to scale the magnitude of contrastive loss on sample $i$ and the corresponding entries in the normalizing factor for each $a_{i, j}$. We ignore comparisons of the same sample with the following modifications $M = M \odot (1 - \mathbb{I})$, $A = A \odot (1 - \mathbb{I})$ where $\odot$ is element-wise multiplication and $\mathbb{I}$ is the identity matrix. We can now define $\mathcal{L}_{\text{SSC}}$,

\begin{equation}
\label{ssc}
\mathcal{L}_{\text{SSC}} = -\frac{1}{n}\sum_{i\in \mathcal{B}} \frac{\omega_i}{\sum m_{i, \cdot}}\sum_{j\in \mathcal{B}} m_{i, j} \log\left(\frac{a_{i, j}}{a_{i, \cdot} \cdot \omega}\right)
\end{equation}

SimCLR and SupCon loss are special cases of $\mathcal{L}_{\text{SSC}}$. In $\mathcal{L}_{\text{SSC}}$, positive pairs are not specified discretely. Every pair of samples has a score in $M$ between 0 and 1 which corresponds to how much evidence there is that they constitute a positive pair. If all data is labeled and all $\omega_i = 1$, $\mathcal{L}_{\text{SSC}}$ becomes SupCon loss. If all data is unlabeled, only transformed pairs have a maximum positive pair relationship, and all $\omega_i = 1$, $\mathcal{L}_{\text{SSC}}$ becomes SimCLR loss. Circumventing the discrete selection of positive pairs is simpler and better reflects the true state of pseudo-labels compared to employing a tunable decision threshold in \cite{Zhang2022SemisupervisedCL} and others.

When training a classifier, we use an Xent loss similar to SEAL loss except that we feed in $q_i$ instead of softmax label vectors. We use the output from $f_g$ instead of $f_z$. We send each $g_i$ through a softmax, $\sigma(g_i)$, and compute,

\begin{equation}
\label{ce}
\mathcal{L}_{\text{CLS}} = -\frac{1}{n} \sum_{i\in \mathcal{B}} \sum_{k\in c} q_{i, k} \log\left( \sigma(g_i)_k \right) 
\end{equation}

One can interpret $\mathcal{L}_{\text{CLS}}$ as a credibility-weighted version of categorical Xent. Recall each $q_i$ will contain only one non-zero entry. Assume this non-zero entry is index $k$. The value of $\mathcal{L}_{\text{CLS}}$ for sample $i$ is $\omega_i \log\left( \sigma(g_i)_k \right)$.

\subsection{The CCP Iteration}
\label{subsec:the_ccp_iteration}

We formalize the CCP iteration in \cref{ccp}, the propagation mechanism in \cref{q_j}, and our subsampling procedure in \cref{kl_div_subsamp_algo}. Each assumes $\{q_l\}_{l\in L}$ are trusted labels. If this is not true, \textit{e.g.} in the noisy-label scenario, all operations applied to unlabeled data are applied to all data. During a single CCP iteration, we continuously predict new $q_i$'s for samples while minimizing $\mathcal{L}_{\text{SSC}}$. We store all credibility vectors and then average them together at the end of the iteration. The SEAL algorithm \cite{https://doi.org/10.48550/arxiv.2012.05458}, bearing similarity to CCP's outermost iteration, is designed to correct the label of mislabeled samples in supervised problems. It was found that, during training, network predictions frequently oscillate between incorrect and correct labels in the presence of label noise. This suggests averaging the predictions made across epochs to correct label noise. Similarly, propagated pseudo-labels are subjected to the randomness of the instantaneous network state \textit{and} batch selection. In the presence of pseudo-label error, and, more generally, uncertainty arising from within the model or data, we observe the same oscillatory behavior reported in \cite{https://doi.org/10.48550/arxiv.2012.05458} of predicted pseudo-labels (shown in \cref{sec:pseudo_label_oscillation}). This grants us a similar theoretical motivation for averaging across epochs in CCP. Assume there is a latent, preferable pseudo-label for sample $x_i$ denoted $q_i^{*}\in \mathbb{R}^K$. Denote the pseudo-label obtained at the end of the $m$-th CCP iteration as $q_i^{[\cdot,m]}$. Based on the oscillatory behavior of pseudo-labels, we can approximate the pseudo-label at the $\xi$-th epoch of the $m$-th iteration for $m\geq 1$ as,

\begin{equation}
\label{q_delta_approximation}
q_i^{[\xi, m]} \approx \alpha^{[\xi, m]}_i \beta^{[\xi, m]}_i + (1-\alpha_i^{[\xi, m]})q_i^{[\cdot,m-1]}
\end{equation}

Where $\xi \in \{1, 2, \ldots, \Xi\}$, $q_i^{[\cdot,0]}$ are zero vectors, $\alpha^{[\xi, m]}_i \in [0, 1]$ are coefficients dependent on instances, the network, and $q_i^{[\cdot,m-1]}$ with $\alpha^{[\xi, 1]}_i=1 \quad \forall \xi$, and $\beta^{[\xi, m]}_i \in \mathbb{R}^K$ are i.i.d. random vectors with $\mathbb{E}[\beta^{[\xi,m]}_i] = q_i^{*}$. Consider a uniformly chosen random epoch and iteration denoted $\xi'$ and $m'$, respectively. We can see CCP is expected to improve the quality of pseudo-labels iteratively and new pseudo-labels at the end of an iteration have lower variance due to averaging.

\begin{equation}
\label{better_than_before}
\lVert \mathbb{E}[q_i^{[\cdot, m'+1]}] - q_i^{*} \rVert \leq \lVert q_i^{[\cdot, m']} - q_i^{*} \rVert
\end{equation}
\begin{equation}
\label{lower_variance}
\text{var}(q_{i,k}^{[\cdot, m']}) \leq \text{var}(q_{i,k}^{[\xi', m']}) \quad \forall k\in c
\end{equation}

\begin{algorithm}[tb]
\caption{An iteration of the CCP algorithm.}
\label{ccp}
\begin{algorithmic}[1]
\State \textbf{Given} $\Xi$, $\mathcal{T}$, $\{q_l\}_{l\in L}$, $X$, $p_{\textrm{last}}$, $d_{\textrm{max}}$
\If{$\{q_u\}_{u\in U}$ are not available}
    \State Initialize $q_u = \vec{0}$ for $u\in U$ \label{ccp:init_q}
\EndIf
\State Reset $f_b$, $f_z$ to a random or pretrained state\label{ccp:init_f_b}
\For{$\xi \in \{1, 2, \ldots, \Xi\}$} \label{ccp:outermost_iter}
    \For{balanced batches $\{(x_i, q_i)\}_{i\in \mathcal{B}_l \cup \mathcal{B}_u}$} \label{ccp:batch_draw}
        \State Form $\{(x_i, q_i), (x_{i+n}, q_i)\}_{i\in \mathcal{B}_l \cup \mathcal{B}_u}$ with randomly drawn $t_1, t_2 \in \mathcal{T}$
        \State Compute $z_i = f_z(f_b(x_i))$, $z_{i+n} = f_z(f_b(x_{i+n}))$ for $i\in \mathcal{B}_l \cup \mathcal{B}_u$
        \State Compute and store $\{\tilde{q}_j\}_{{j}\in \mathcal{B}_u}$ using \cref{q_j}
        \State Train $f_b$, $f_z$ using the gradient of $\mathcal{L}_{\text{SSC}}$ computed with $\{(z_i, q_i), (z_{i+n}, q_i)\}_{i\in \mathcal{B}_l \cup \mathcal{B}_u}$
    \EndFor
\EndFor
 
\State Compute $\{\hat{q}_u =$ average of all stored $\tilde{q}_u\}_{u \in U}$ \label{ccp:avg_over_epoch}
\State Scale all $\hat{q}_u$ via multiplying by $\frac{1}{\gamma}$ \label{ccp:scale}
\For{$k\in c$} \label{ccp:begin_cred_adj} \Comment{Credibility adjustment}
    \State $q_{u, k} = \hat{q}_{u, k} - \max_{k'\in c\setminus k}(\hat{q}_{u, k'})$ for $u\in U$ \label{ccp:cred_adj}
\EndFor \label{ccp:end_cred_adj}
\State Clip all values in $\{q_u\}_{u\in U}$ to lie in $[0,1]$ \label{ccp:clip}
\State Compute $p$, $\{\hat{\omega}_u\}_{u\in U}$ using \cref{kl_div_subsamp_algo} \label{ccp:compute_p}
\State Reset bottom $p\%$ of $\{q_u\}_{u\in U}$ ordered by $\{\hat{\omega}_u\}_{u\in U}$ \label{ccp:subsamp}
\State Update $p_{\textrm{last}} = p$, $d_{\textrm{max}} = \nicefrac{d_{\textrm{max}}}{2}$ \label{ccp:update_p_last}
\State \textbf{Return} $f_b$, $p_{\textrm{last}}$, $d_{\textrm{max}}$, $\{q_u\}_{u\in U}$ (used in next iteration) \label{ccp:return}
\end{algorithmic}
\end{algorithm}

\begin{algorithm}[tb]
\caption{Compute propagated credibility vectors.}
\label{q_j}
\begin{algorithmic}[1]
\State \textbf{Given} $\{(z_i, q_i), (z_{i+n}, q_i)\}_{i\in \mathcal{B}_l \cup \mathcal{B}_u}$
\State Define expanded indices $\mathcal{B}' = \{0, 1, \ldots, 2n-1\}$
\State Define $q_{i+n} = q_i$ for $i\in \mathcal{B}$
\For{$j \in \mathcal{B}_u$}
    \For{$k \in c$} \Comment{Compute class similarities}
        \State $\psi_{j, k} = \nicefrac{\sum_{i}^{\mathcal{B}'\setminus j} \phi(z_{j}, z_i)q_{i,k}}{\sum_{i}^{\mathcal{B}'\setminus j} q_{i,k}}$
        \State $\psi_{j+n, k} = \nicefrac{\sum_{i}^{\mathcal{B}'\setminus j+n} \phi(z_{j+n}, z_i)q_{i,k}}{\sum_{i}^{\mathcal{B}'\setminus j+n} q_{i,k}}$
    \EndFor
    \For{$k \in c$} \label{q_j:begin_cred_adj} \Comment{Credibility adjustment}
        \State $\tilde{q}_{j, k} = \psi_{j, k} - \max_{k'\in c\setminus k}(\psi_{j, k'})$ \label{q_j:cred_adj}
        \State $\tilde{q}_{j+n, k} = \psi_{j+n, k} - \max_{k'\in c\setminus k}(\psi_{j+n, k'})$
    \EndFor \label{q_j:end_cred_adj}
    \State Store $\tilde{q}_j = \frac{\tilde{q}_j + \tilde{q}_{j+n}}{2}$ \label{q_j:avg_over_trans} \Comment{Average across $t_1$, $t_2$}
\EndFor
\State \textbf{Return} $\{\tilde{q}_j\}_{j\in \mathcal{B}_u}$
\end{algorithmic}
\end{algorithm}

In \cref{ccp}, \cref{ccp:avg_over_epoch,ccp:scale,ccp:begin_cred_adj,ccp:cred_adj,ccp:end_cred_adj,ccp:clip}, we adjust labels in several ways to make them more suitable as labels. In \cref{ccp:scale}, we multiply by a scaling factor $\nicefrac{1}{\gamma}$ after the average. We use $\gamma=\max_{u\in U, c\in K} \hat{q}_{u, c}$. This ensures the strongest pseudo-label will have a strength of $1$. Before we clip all values outside the range of $[0,1]$ in \cref{ccp:clip}, we compute a final credibility adjustment. The scaled, credibility adjusted, and clipped vectors are denoted $\{q_u\}_{u\in U}$. In \cref{ccp:batch_draw}, we find more stable performance when using \textit{balanced} batches during CCP iterations. We guarantee there are $\ge 1$ (pseudo-)labeled samples per class in every batch via oversampling. Pseudo-labels obtained from repeated uses of a sample are simply gathered together in the average. This ensures 1) the correct class can be propagated and 2) credibility can be properly measured. Lastly, we found it beneficial to use a small learning rate in the first iteration. Especially for few-label scenarios, the network quickly overfits to the small labeled set, and pseudo-label quality suffers. Specifically, we reduced the original learning rate of $0.06$ to $0.0006$ in the first iteration for all experiments.

\subsubsection{$D_{\text{KL}}$-Based Subsampling}
\label{subsubsec:subsampling}

We use \cref{kl_div_subsamp_algo} to decide which pseudo-labels to reset to $\vec{0}$ between CCP iterations. This is inspired by self-training \cite{https://doi.org/10.48550/arxiv.2202.12040} which solves the similar but inverted task of iteratively \textit{adding} data to the labeled set. These techniques typically use the max score of a learned classifier as an indicator of confidence. Similarly, our confidence indicator for a credibility vector is the maximum of its \textit{unclipped} values. We compute a percentage $p \in \{0\%, 1\%, \ldots, 99\%\}$ that represents what percent of least confident pseudo-labels will be reset. Aggressive $p$ values can cause instability or slow convergence when too many correct pseudo-labels are reset. We hypothesize the instability is similar in cause to the instability of self-training mechanisms \cite{https://doi.org/10.48550/arxiv.2202.07136}. CCP is shown highly stable without any subsampling (\cref{subsec:ablation_analysis}). Accordingly, our approach is to balance resetting weak pseudo-labels with limiting the divergence of the predicted class distribution of the selected unlabeled data from the predicted class distribution of all unlabeled data. Concretely, in \cref{kl_div_subsamp_algo:q} we compute the anchor distribution, $Q$, by summing the weights for every class and dividing by the total mass. We search over candidate $p$'s with one minus the $p$ used in the previous iteration, $p_{\text{last}} - 1\%$, as the maximum to ensure we don't increase $p$ between iterations. At \cref{kl_div_subsamp_algo:p}, we compute the new distribution, $P$, obtained after resetting the candidate percentage of vectors. At \cref{kl_div_subsamp_algo:d}, we compute the Kullback–Leibler (KL) divergence \cite{10.1214/aop/1176996454} between these distributions. At \cref{kl_div_subsamp_algo:choose_p}, we choose $p$ by selecting the maximum candidate that obeys a strict limit on the divergence, $d_{\textrm{max}}$. We divide $d_{\textrm{max}}$ by $2$ for the next iteration to support convergence. Our method is unsupervised, free of imposing assumptions, and normalized to the dataset size. This suggests, in theory, that a single schedule for $d_{\textrm{max}}$ should generalize well across datasets. Indeed, we find an initial value of $d_{\textrm{max}}=0.01$ to work \textit{generally} well across all of our experiments.

\begin{algorithm}[tb]
\caption{Compute a subsample percentage.}
\label{kl_div_subsamp_algo}
\begin{algorithmic}[1]
\State \textbf{Given} $\{\hat{q}_u\}_{u\in U}$, $\{q_u\}_{u\in U}$, $p_{\textrm{last}}$, $d_{\textrm{max}}$
\State Compute $\{\hat{\omega}_u = \max(\hat{q}_u)\}_{u\in U}$ \label{kl_div_subsamp_algo:omega}
\State $Q = \nicefrac{\sum_u^U q_u}{\sum_k^c \sum_u^U q_{u,k}}$ \Comment{Anchor distribution} \label{kl_div_subsamp_algo:q}
\For{$p_i \in \{0\%, 1\%, \ldots, p_{\textrm{last}}-1\%\}$}
    \State Reset bottom $p_i\%$ of $\{q_u\}_{u\in U}$ ordered by $\{\hat{\omega}_u\}_{u\in U}$
    \State $P = \nicefrac{\sum_u^U q_u}{\sum_k^c \sum_u^U q_{u,k}}$ \Comment{Candidate distribution} \label{kl_div_subsamp_algo:p}
    \State $d_i = D_{\text{KL}}(P \parallel Q) = \sum_{k\in c} P_k\log_2\left(\nicefrac{P_k}{Q_k}\right)$ \label{kl_div_subsamp_algo:d}
\EndFor
\State $p=\max(\{p_i \text{ for all } i \text{ such that } d_i<d_{\textrm{max}}\})$ \label{kl_div_subsamp_algo:choose_p}
\State \textbf{Return} $p$, $\{\hat{\omega}_u\}_{u\in U}$
\end{algorithmic}
\end{algorithm}

\subsection{Building a Classifier}
\label{subsec:building_a_classifier}

After CCP iterations have concluded, we apply the soft labels, $\{q_j\}_{j\in I}$, to build a classifier consisting of $f_b$, $f_g$. We use Stochastic Gradient Descent with randomly selected mini-batches (regardless of whether the data was originally labeled). We find that minimizing both \cref{ssc} (using $f_z$) and \cref{ce} together consistently outperforms \cref{ce} alone. In our experiments, resetting $f_b$, $f_z$ (to a random or pretrained state) after the final iteration of CCP is marginally more performant than reusing the state after the last iteration -- this prior state will have been learned without the latest pseudo-label adjustments. However, the latter converges much faster. We find that using \cref{kl_div_subsamp_algo} to subsample the final $\{q_j\}_{j\in I}$ greatly increases training set accuracy but has a minimal and inconsistent effect on the test set. This is because cropping the \say{hard} samples with weak pseudo-labels at this stage makes the training set easier but reduces generalization.

\section{Experimental Results\footnote{Code available at: \url{https://github.com/PaloAltoNetworks/CCP_CIFAR} and \url{https://github.com/PaloAltoNetworks/ccp-as-pytorch}}}
\label{sec:experimental_results}

We leverage CIFAR-10 and CIFAR-100 for our data variable sensitivity experiments \cite{Krizhevsky2009a}. We use these datasets as starting points to explore the five data variables we introduce in \cref{sec:intro}. For comparability among the data variables, we define a base case that represents a minimal severity of each data variable. We study each data variable independently by increasing the severity from the base case at three levels. For the base cases, we take the first $40\%$ of classes and define them as in-distribution (ID) \text{e.g.} for CIFAR-10, classes $[0, 3]$ are ID while $[4, 9]$ are out-of-distribution (OOD). For CIFAR-10, the base case is defined as 400 labeled and 4600 unlabeled samples per ID class. For CIFAR-100, we have 100 labeled and 400 unlabeled samples per ID class. We then independently introduce the following perturbations:

\begin{enumerate}
    \item \textbf{Few-label}: We move labeled data to the unlabeled set such that the number of labeled samples per ID class reduces to 25, 4, and 2. 
    \item \textbf{Open-set}: We move all data from OOD classes into the unlabeled set incrementally \textit{e.g.} on CIFAR-10 the unlabeled set contains data from classes $[0, 5]$, $[0, 7]$, and $[0, 9]$.
    \item \textbf{Noisy-label}: We randomly select $20\%$, $40\%$, and $60\%$ of labels and randomly perturb them to an incorrect ID class.
    \item \textbf{Class distribution imbalance/misalignment}: We perturb labeled and unlabeled sets separately. We take the last $50\%$ of ID classes, \textit{e.g.} for CIFAR-100 the ID classes we perturb are $[20, 39]$, and reduce their quantity (discarding samples). When perturbing unlabeled sets, we reduce to $20\%$, $10\%$, and $0\%$ of the original quantity. For labeled sets, we reduce such that 25, 4, and 2 samples remain.
\end{enumerate}


We perform our data variable sensitivity analysis on CCP, CoMatch \cite{Li2021CoMatchSL}, OpenMatch \cite{NEURIPS2021_da11e8cd}, ACR \cite{Wei_2023_CVPR}, and FixMatch \cite{https://doi.org/10.48550/arxiv.2001.07685} with and without distribution alignment. CoMatch and FixMatch were developed to optimize performance in the few-label scenario, whereas OpenMatch and ACR were designed to address open-set and misalignment in the unlabeled set, respectively. We used the original author's implementations for all algorithms. For all algorithms and datasets, we use the standard WRN-28-2 and WRN-28-8 \cite{DBLP:journals/corr/ZagoruykoK16} as the backbone network architecture for CIFAR-10 and CIFAR-100, respectively. We closely match all training hyperparameters and settings of \cite{https://doi.org/10.48550/arxiv.2001.07685,Li2021CoMatchSL}. For algorithm-specific hyperparameters, we use the values originally recommended in the work for the corresponding dataset. CCP-specific hyperparameter choices are detailed in \cref{tab:ccp_hyperparameters}. Some minor differences exist across all algorithm implementations \textit{e.g.} the parameter choices for each $t\in\mathcal{T}$, the deep learning software package used, and regularization used. We report the performance of a fully supervised control for each algorithm implementation to help understand the effect. 


\subsubsection{Computational Expense}
\label{subsubsec:computational_expense}

CCP iterations can potentially be computationally taxing at large values of $\Xi$. In the worst case, one must fully train a new network for every CCP iteration. Additionally, if $\Xi$ is too large, incorrect pseudo-labels are memorized and $\hat{q}$ becomes biased towards the error. We found beneficial a course of pretraining using $\mathcal{L}_{\text{SSC}}$'s unsupervised counterpart, SimCLR, to determine the initialization of $f_b$, $f_z$. This pretrained state significantly reduced the inherent randomness of early pseudo-label predictions. Also, due to faster convergence, using this pretrained state allowed us to use aggressive early stopping after the first CCP iteration which has a fixed number of epochs and a low learning rate (50 and 0.0006, respectively). Specifically, we maintain a batch-wise exponential moving average (EMA) of $\mathcal{L}_{\text{SSC}}$ with decay 0.99 during each CCP iteration and halt quickly after the EMA stops dropping. This resulted in the equivalent of $\sim 15$ and $\sim 50$ epochs per iteration with 24 and 12 CCP iterations for CIFAR-10 and CIFAR-100, respectively. However, shown in \cref{subsec:ablation_analysis}, CCP often converges after only a few iterations. The only deviation from this is in the open-set experiments, in which we report the performance after a single CCP iteration. Further iterations increased pseudo-label confidence in both ID and OOD samples and thus did not provide value. Using \cref{kl_div_subsamp_algo} with a more aggressive $d_{\textrm{max}}$ and schedule mitigated this problem however we found satisfying results simply with a single CCP iteration which was more consistent with the other experiments.

\subsection{Data Variable Sensitivity Analysis}
\label{subsec:dataset_variable_sensitivity_analysis}

We report the test accuracy of each algorithm on each data variable experiment at all levels of severity in \cref{tab:data_var_sensitivity}. We find relatively consistent performance across the control experiments. Performance on the base case reflects an algorithm's inherent label efficiency (which impacts performance in all scenarios) and helps to contextualize all results in this table. We also normalized results to the base case to help visualize an algorithm's sensitivity to each data variable in isolation. These plots can be found in \cref{sec:normalized_data_variable_sensitivity_analysis}. Although CCP doesn't achieve superiority in every scenario, particularly concerning label efficiency, it demonstrates remarkable consistency. For instance, on CIFAR-10, its accuracy never drops below $90\%$ for any scenario representing only a $5\%$ drop from the base case. Moreover, CCP uniquely outperforms a supervised baseline in every experiment. This supports the core reliability thesis. Where CCP does not achieve superiority, label efficiency seems to be a large contributing factor. Other algorithms fail, sometimes catastrophically below CCP or even the supervised baseline, in the presence of certain data variations. This is most prevalent when perturbing the labeled distribution and the relatively poorly explored noisy-label scenario where CCP often outperforms by a large margin. Intuitively, DA seems to help FixMatch marginally in the few-label scenario but hurts when perturbing the unlabeled data distribution. Note that all algorithms perform relatively well on open-set experiments. All algorithms tested besides CCP build off FixMatch in some way. This is common practice in the literature. We hypothesize that, for these algorithms, FixMatch's pseudo-label threshold parameter effectively removes unconfident pseudo-labeled OOD samples from consideration. For CCP, credibility adjustments shrink the pseudo-labels of OOD samples to near zero automatically.

\begin{table*}[tb]
\centering
\fontsize{10}{11}\selectfont
\fontfamily{ptm}\selectfont
\begin{tabular}{@{}clclclclclclc@{}}
\toprule
\textbf{CIFAR-10} &
  \multirow{2}{*}{} &
  \textbf{Control} &
  \multirow{2}{*}{} &
  \multirow{2}{*}{\textbf{Few-label}} &
  \multirow{2}{*}{} &
  \multirow{2}{*}{\textbf{Open-set}} &
  \multirow{2}{*}{} &
  \multirow{2}{*}{\textbf{Noisy-label}} &
  \multirow{2}{*}{} &
  \multicolumn{3}{c}{\textbf{Class Imbalance/Misalignment}} \\ \cmidrule(r){1-1} \cmidrule(l){11-13} 
\textbf{CIFAR-100} &
   &
  \multicolumn{1}{l}{\textbf{Base Case}} &
   &
   &
   &
   &
   &
   &
   &
  $U$ &
   &
  $L$ \\ \midrule
\multirow{2}{*}{\begin{tabular}[c]{@{}c@{}}Supervised\\ Baseline\end{tabular}} &
  \multirow{2}{*}{} &
  \begin{tabular}[c]{@{}c@{}}97.00\%\\ 89.18\%\end{tabular} &
  \multirow{2}{*}{} &
  \begin{tabular}[c]{@{}c@{}}63.33\%\\ 54.93\%\\ 49.98\%\end{tabular} &
  \multirow{2}{*}{} &
  \begin{tabular}[c]{@{}c@{}}89.18\%\\ 89.18\%\\ 89.18\%\end{tabular} &
  \multirow{2}{*}{} &
  \begin{tabular}[c]{@{}c@{}}78.83\%\\ 73.08\%\\ 58.60\%\end{tabular} &
  \multirow{2}{*}{} &
  \begin{tabular}[c]{@{}c@{}}89.40\%\\ 89.40\%\\ 89.40\%\end{tabular} &
  \multirow{2}{*}{} &
  \begin{tabular}[c]{@{}c@{}}75.90\%\\ 63.98\%\\ 58.08\%\end{tabular} \\ \cmidrule(lr){3-3} \cmidrule(lr){5-5} \cmidrule(lr){7-7} \cmidrule(lr){9-9} \cmidrule(lr){11-11} \cmidrule(l){13-13} 
 &
   &
  \begin{tabular}[c]{@{}c@{}}84.10\%\\ 67.30\%\end{tabular} &
   &
  \begin{tabular}[c]{@{}c@{}}43.85\%\\ 20.03\%\\ 15.38\%\end{tabular} &
   &
  \begin{tabular}[c]{@{}c@{}}67.30\%\\ 67.30\%\\ 67.30\%\end{tabular} &
   &
  \begin{tabular}[c]{@{}c@{}}55.45\%\\ 47.35\%\\ 35.70\%\end{tabular} &
   &
  \begin{tabular}[c]{@{}c@{}}67.30\%\\ 67.30\%\\ 67.30\%\end{tabular} &
   &
  \begin{tabular}[c]{@{}c@{}}54.03\%\\ 38.95\%\\ 36.13\%\end{tabular} \\
\multicolumn{13}{c}{} \\
\multirow{2}{*}{CoMatch} &
  \multirow{2}{*}{} &
  \begin{tabular}[c]{@{}c@{}}97.00\%\\ 97.74\%\end{tabular} &
  \multirow{2}{*}{} &
  \begin{tabular}[c]{@{}c@{}}96.97\%\\ 97.45\%\\ 97.17\%\end{tabular} &
  \multirow{2}{*}{} &
  \begin{tabular}[c]{@{}c@{}}97.82\%\\ 97.69\%\\ 97.79\%\end{tabular} &
  \multirow{2}{*}{} &
  \begin{tabular}[c]{@{}c@{}}91.82\%\\ \textbf{68.69\%}\\ \textbf{50.05\%}\end{tabular} &
  \multirow{2}{*}{} &
  \begin{tabular}[c]{@{}c@{}}94.07\%\\ 91.54\%\\ \textbf{87.77\%}\end{tabular} &
  \multirow{2}{*}{} &
  \begin{tabular}[c]{@{}c@{}}96.11\%\\ 89.73\%\\ 90.90\%\end{tabular} \\ \cmidrule(lr){3-3} \cmidrule(lr){5-5} \cmidrule(lr){7-7} \cmidrule(lr){9-9} \cmidrule(lr){11-11} \cmidrule(l){13-13} 
 &
   &
  \begin{tabular}[c]{@{}c@{}}85.54\%\\ 85.02\%\end{tabular} &
   &
  \begin{tabular}[c]{@{}c@{}}82.12\%\\ 66.02\%\\ 61.51\%\end{tabular} &
   &
  \begin{tabular}[c]{@{}c@{}}84.90\%\\ 84.72\%\\ 85.24\%\end{tabular} &
   &
  \begin{tabular}[c]{@{}c@{}}78.72\%\\ 67.16\%\\ 47.94\%\end{tabular} &
   &
  \begin{tabular}[c]{@{}c@{}}80.23\%\\ 78.52\%\\ 77.75\%\end{tabular} &
   &
  \begin{tabular}[c]{@{}c@{}}81.00\%\\ 61.41\%\\ 48.04\%\end{tabular} \\
\multicolumn{13}{l}{} \\
\multirow{2}{*}{ACR} &
  \multirow{2}{*}{} &
  \begin{tabular}[c]{@{}c@{}}97.38\%\\ 97.20\%\end{tabular} &
  \multirow{2}{*}{} &
  \begin{tabular}[c]{@{}c@{}}95.45\%\\ 67.66.\%\\ 60.66\%\end{tabular} &
  \multirow{2}{*}{} &
  \begin{tabular}[c]{@{}c@{}}96.80\%\\ 96.42\%\\ 96.78\%\end{tabular} &
  \multirow{2}{*}{} &
  \begin{tabular}[c]{@{}c@{}}82.90\%\\ \textbf{62.73\%}\\ \textbf{39.75\%}\end{tabular} &
  \multirow{2}{*}{} &
  \begin{tabular}[c]{@{}c@{}}93.90\%\\ 94.45\%\\ \textbf{88.55\%}\end{tabular} &
  \multirow{2}{*}{} &
  \begin{tabular}[c]{@{}c@{}}96.05\%\\ 94.85\%\\ 87.75\%\end{tabular} \\ \cmidrule(lr){3-3} \cmidrule(lr){5-5} \cmidrule(lr){7-7} \cmidrule(lr){9-9} \cmidrule(lr){11-11} \cmidrule(l){13-13} 
 &
   &
  \begin{tabular}[c]{@{}c@{}}85.70\%\\ 82.80\%\end{tabular} &
   &
  \begin{tabular}[c]{@{}c@{}}74.97\%\\ 42.83\%\\ 22.07\%\end{tabular} &
   &
  \begin{tabular}[c]{@{}c@{}}82.78\%\\ 82.88\%\\ 82.33\%\end{tabular} &
   &
  \begin{tabular}[c]{@{}c@{}}72.70\%\\ 56.33\%\\ \textbf{33.38\%}\end{tabular} &
   &
  \begin{tabular}[c]{@{}c@{}}78.67\%\\ 76.95\%\\ 75.80\%\end{tabular} &
   &
  \begin{tabular}[c]{@{}c@{}}77.70\%\\ 64.30\%\\ 52.65\%\end{tabular} \\
\multicolumn{13}{l}{} \\
\multirow{2}{*}{OpenMatch} &
  \multirow{2}{*}{} &
  \begin{tabular}[c]{@{}c@{}}96.80\%\\ 96.65\%\end{tabular} &
  \multirow{2}{*}{} &
  \begin{tabular}[c]{@{}c@{}}92.20\%\\ 68.18\%\\ \textbf{43.08\%}\end{tabular} &
  \multirow{2}{*}{} &
  \begin{tabular}[c]{@{}c@{}}93.58\%\\ 91.93\%\\ 91.00\%\end{tabular} &
  \multirow{2}{*}{} &
  \begin{tabular}[c]{@{}c@{}}91.28\%\\ 85.93\%\\ 76.75\%\end{tabular} &
  \multirow{2}{*}{} &
  \begin{tabular}[c]{@{}c@{}}94.90\%\\ 93.93\%\\ 91.48\%\end{tabular} &
  \multirow{2}{*}{} &
  \begin{tabular}[c]{@{}c@{}}76.43\%\\ \textbf{50.55\%}\\ \textbf{49.80\%}\end{tabular} \\ \cmidrule(lr){3-3} \cmidrule(lr){5-5} \cmidrule(lr){7-7} \cmidrule(lr){9-9} \cmidrule(lr){11-11} \cmidrule(l){13-13} 
 &
   &
  \begin{tabular}[c]{@{}c@{}}85.20\%\\ 83.68\%\end{tabular} &
   &
  \begin{tabular}[c]{@{}c@{}}77.30\%\\ 40.35\%\\ 27.88\%\end{tabular} &
   &
  \begin{tabular}[c]{@{}c@{}}83.13\%\\ 82.63\%\\ 82.33\%\end{tabular} &
   &
  \begin{tabular}[c]{@{}c@{}}77.35\%\\ 67.10\%\\ 53.55\%\end{tabular} &
   &
  \begin{tabular}[c]{@{}c@{}}80.80\%\\ 79.18\%\\ 76.05\%\end{tabular} &
   &
  \begin{tabular}[c]{@{}c@{}}79.55\%\\ 54.95\%\\ 49.15\%\end{tabular} \\
\multicolumn{13}{l}{} \\
\multirow{2}{*}{\begin{tabular}[c]{@{}c@{}}FixMatch\\ w/o DA\end{tabular}} &
  \multirow{2}{*}{} &
  \begin{tabular}[c]{@{}c@{}}97.20\%\\ 97.97\%\end{tabular} &
  \multirow{2}{*}{} &
  \begin{tabular}[c]{@{}c@{}}97.37\%\\ 96.35\%\\ 76.26\%\end{tabular} &
  \multirow{2}{*}{} &
  \begin{tabular}[c]{@{}c@{}}97.59\%\\ 97.52\%\\ 97.67\%\end{tabular} &
  \multirow{2}{*}{} &
  \begin{tabular}[c]{@{}c@{}}91.57\%\\ \textbf{66.44\%}\\ \textbf{44.62\%}\end{tabular} &
  \multirow{2}{*}{} &
  \begin{tabular}[c]{@{}c@{}}95.76\%\\ 93.82\%\\ 90.67\%\end{tabular} &
  \multirow{2}{*}{} &
  \begin{tabular}[c]{@{}c@{}}96.11\%\\ 68.27\%\\ \textbf{54.40\%}\end{tabular} \\ \cmidrule(lr){3-3} \cmidrule(lr){5-5} \cmidrule(lr){7-7} \cmidrule(lr){9-9} \cmidrule(lr){11-11} \cmidrule(l){13-13} 
 &
   &
  \begin{tabular}[c]{@{}c@{}}85.44\%\\ 85.66\%\end{tabular} &
   &
  \begin{tabular}[c]{@{}c@{}}80.80\%\\ 56.82\%\\ 42.39\%\end{tabular} &
   &
  \begin{tabular}[c]{@{}c@{}}85.09\%\\ 85.02\%\\ 85.24\%\end{tabular} &
   &
  \begin{tabular}[c]{@{}c@{}}75.42\%\\ 62.43\%\\ 41.84\%\end{tabular} &
   &
  \begin{tabular}[c]{@{}c@{}}80.73\%\\ 78.57\%\\ 76.41\%\end{tabular} &
   &
  \begin{tabular}[c]{@{}c@{}}80.01\%\\ 60.32\%\\ 53.00\%\end{tabular} \\
\multicolumn{13}{l}{} \\
\multirow{2}{*}{\begin{tabular}[c]{@{}c@{}}FixMatch\\ w/ DA\end{tabular}} &
  \multirow{2}{*}{} &
  \begin{tabular}[c]{@{}c@{}}96.80\%\\ 98.12\%\end{tabular} &
  \multirow{2}{*}{} &
  \begin{tabular}[c]{@{}c@{}}97.12\%\\ 97.12\%\\ 96.80\%\end{tabular} &
  \multirow{2}{*}{} &
  \begin{tabular}[c]{@{}c@{}}97.52\%\\ 97.62\%\\ 97.30\%\end{tabular} &
  \multirow{2}{*}{} &
  \begin{tabular}[c]{@{}c@{}}91.12\%\\ \textbf{65.97\%}\\ \textbf{46.97\%}\end{tabular} &
  \multirow{2}{*}{} &
  \begin{tabular}[c]{@{}c@{}}93.58\%\\ 91.57\%\\ \textbf{88.44\%}\end{tabular} &
  \multirow{2}{*}{} &
  \begin{tabular}[c]{@{}c@{}}97.40\%\\ 92.65\%\\ 80.37\%\end{tabular} \\ \cmidrule(lr){3-3} \cmidrule(lr){5-5} \cmidrule(lr){7-7} \cmidrule(lr){9-9} \cmidrule(lr){11-11} \cmidrule(l){13-13} 
 &
   &
  \begin{tabular}[c]{@{}c@{}}85.59\%\\ 85.29\%\end{tabular} &
   &
  \begin{tabular}[c]{@{}c@{}}81.40\%\\ 58.09\%\\ 43.23\%\end{tabular} &
   &
  \begin{tabular}[c]{@{}c@{}}84.97\%\\ 84.42\%\\ 84.75\%\end{tabular} &
   &
  \begin{tabular}[c]{@{}c@{}}75.97\%\\ 61.81\%\\ 40.95\%\end{tabular} &
   &
  \begin{tabular}[c]{@{}c@{}}79.32\%\\ 77.21\%\\ 75.15\%\end{tabular} &
   &
  \begin{tabular}[c]{@{}c@{}}81.75\%\\ 69.27\%\\ 63.79\%\end{tabular} \\
\multicolumn{13}{l}{} \\
\multirow{2}{*}{CCP (Ours)} &
  \multirow{2}{*}{} &
  \begin{tabular}[c]{@{}c@{}}96.85\%\\ 95.03\%\end{tabular} &
  \multirow{2}{*}{} &
  \begin{tabular}[c]{@{}c@{}}94.50\%\\ 90.23\%\\ 90.28\%\end{tabular} &
  \multirow{2}{*}{} &
  \begin{tabular}[c]{@{}c@{}}94.83\%\\ 94.50\%\\ 94.40\%\end{tabular} &
  \multirow{2}{*}{} &
  \begin{tabular}[c]{@{}c@{}}94.33\%\\ 94.40\%\\ 94.28\%\end{tabular} &
  \multirow{2}{*}{} &
  \begin{tabular}[c]{@{}c@{}}93.60\%\\ 92.90\%\\ 90.45\%\end{tabular} &
  \multirow{2}{*}{} &
  \begin{tabular}[c]{@{}c@{}}94.45\%\\ 90.28\%\\ 90.25\%\end{tabular} \\ \cmidrule(lr){3-3} \cmidrule(lr){5-5} \cmidrule(lr){7-7} \cmidrule(lr){9-9} \cmidrule(lr){11-11} \cmidrule(l){13-13} 
 &
   &
  \begin{tabular}[c]{@{}c@{}}84.10\%\\ 78.55\%\end{tabular} &
   &
  \begin{tabular}[c]{@{}c@{}}74.05\%\\ 65.38\%\\ 61.38\%\end{tabular} &
   &
  \begin{tabular}[c]{@{}c@{}}78.40\%\\ 77.68\%\\ 77.65\%\end{tabular} &
   &
  \begin{tabular}[c]{@{}c@{}}74.60\%\\ 74.48\%\\ 71.08\%\end{tabular} &
   &
  \begin{tabular}[c]{@{}c@{}}76.25\%\\ 74.63\%\\ 73.70\%\end{tabular} &
   &
  \begin{tabular}[c]{@{}c@{}}75.68\%\\ 69.60\%\\ 65.88\%\end{tabular} \\ \bottomrule
\end{tabular}
\caption{Test set accuracy of each algorithm in each data variable scenario on CIFAR-10 (upper) and CIFAR-100 (lower). In the upper/lower portions of cells, the performance on each severity level is presented in descending order. $U$ ($L$) refers to the perturbation of the unlabeled (labeled) distribution. Bold indicates under-performing the supervised baseline in that scenario.}
\label{tab:data_var_sensitivity}
\end{table*}

\subsection{Ablation Analysis}
\label{subsec:ablation_analysis}

We investigate the criticality of credibility adjustments to successful CCP iterations as well as the effectiveness of subsampling via \cref{kl_div_subsamp_algo}. We focus on the base case and few-label experiments, however, repeating these experiments with the other data variables provided a similar result. 

\begin{figure}[tb]
  \centering
  \includegraphics[scale=0.5]{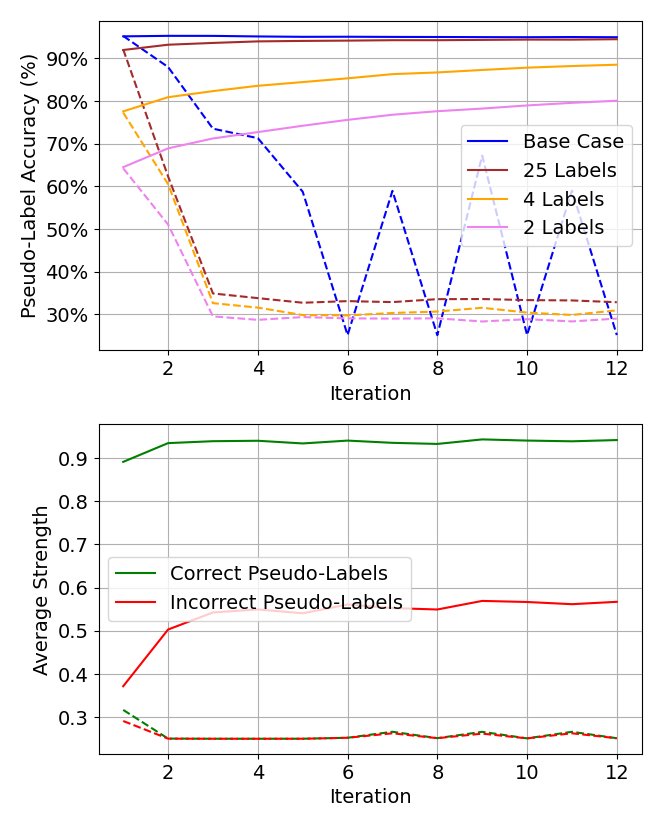}
  \caption{Performance of CCP iterations on the base case and few-label experiments of CIFAR-10 with and without credibility. Solid lines indicate the use of credibility adjustments. Dashed lines indicate a softmax function. \textit{Top:} Pseudo-label accuracy. \textit{Bottom:} The average strength of correct and incorrect pseudo-labels in the base case.}
  \label{cred_ablation_plot}
\end{figure}

In \cref{cred_ablation_plot}, we study the effect of credibility adjustments by measuring the difference in CCP iteration performance when traditional softmax functions replace them. We also omit scaling, clipping, and subsampling when using softmax such that it resembles the SEAL algorithm. We often find quick and catastrophic degradation to maximum entropy pseudo-labels when using softmax. When considering the average strength (maximum value) of pseudo-labels, it is clear that credibility strongly differentiates correct and incorrect pseudo-labels, making it a better fit as a measure of confidence. Recall in \cref{cred_vecs} that, when faced with uncertainty, an Xent gradient with a softmax label pushes the network to produce a high entropy pseudo-label. That mirrors what we see here.

\begin{figure}[tb]
  \centering
  \includegraphics[scale=0.5]{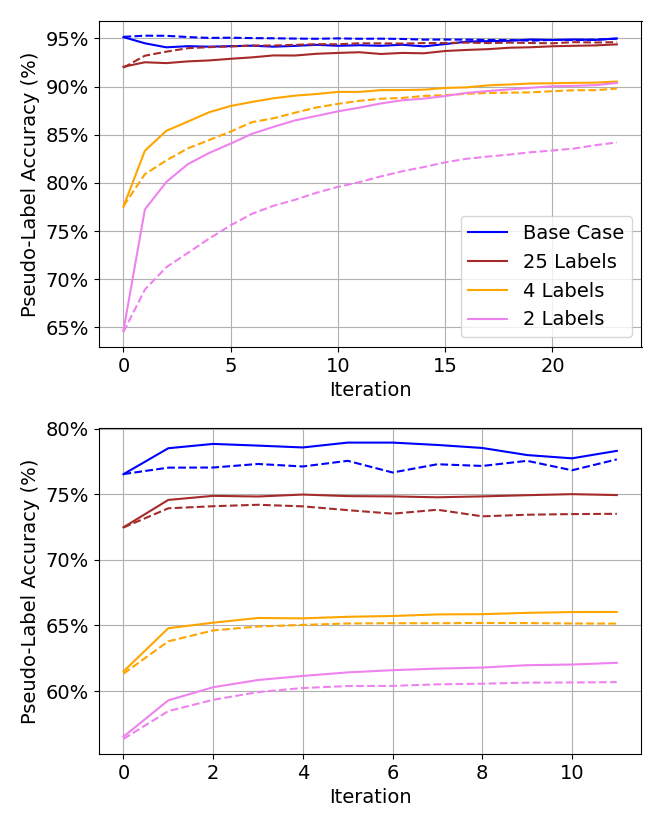}
  \caption{Pseudo-label accuracy during CCP iterations in the base case and few-label experiments of CIFAR-10 (top) and CIFAR-100 (bottom). Solid lines depict the usage of $D_{\text{KL}}$-based subsampling with the parameters and schedule presented in \cref{ccp} with an initial $d_{\textrm{max}}=0.01$. Dashed lines indicate no subsampling \textit{i.e.} $d_{\textrm{max}}=0.0$.}
  \label{kl_ablation_plot}
\end{figure}

In \cref{kl_ablation_plot}, we see $D_{\text{KL}}$-based subsampling provides substantial benefit to the pseudo-label accuracy during CCP iterations at high data variable severity. At worst, it appears to provide no benefit as in the CIFAR-10 base case experiment. Note we used the same initial $d_{\textrm{max}}=0.01$ and schedule presented in \cref{ccp} for all experiments in this work. If instability is observed, which can occur if $d_{\textrm{max}}$ is too large, tuning is necessary. Additional subsampling ablation analysis with a similar result can be found in \cref{tab:ccp_iter_perf_full}.

\section{Conclusion}
\label{sec:conclusion}

We have presented an algorithm that combines a soft contrastive approach to pseudo-labeling with an outer iteration designed for learning under instance-dependent label noise. The result is a highly reliable and effective SSL algorithm that does not perform worse than a supervised baseline across five common real-world SSL scenarios. Future work may include augmenting \cref{q_j} to include successful components from prior work such as consistency training between weak/strong augmentation \cite{https://doi.org/10.48550/arxiv.2001.07685} and instance/semantic similarity \cite{simmatch} to combine CCP's reliability with the label efficiency of other work.

\FloatBarrier

\section*{Acknowledgements}
\label{sec:acknowledgements}

We'd like to thank the many AI experts at Palo Alto Networks for their helpful commentary on this work. In particular, we'd like to thank Sheng Yang and Aaron Isaksen.

\bibliography{main}


\onecolumn
\appendix

\section{CIFAR Training Details}
\label{sec:training_details}

\begin{table}[tb]
\centering
\begin{tabular}{@{}cl@{}}
\textbf{Hyperparameter} & \textbf{Value} \\ \midrule
Batch Size (Warmup/CCP Iterations) & 512 \\
Warmup Epochs & 512 \\
Batch Size (Classifier Training) & 64 \\
Classifier Training Epochs & 512 \\
Learning Rate (First CCP Iteration) & 0.0006 \\
Learning Rate & 0.06 \\
Similarity Metric & Angular \\
$f_z$ output dimension & 128 \\
Activation & ReLU \\
Temperature $\tau$ & 0.01 \\
EMA loss decay & 0.99 \\
EMA model decay & 0.999 \\
$\lambda$ for L2 regularization & 0.0005 \\
Optimizer & SGD with Nesterov Momentum $=0.9$ \\
Learning Rate Schedule & Cosine Decay $\coloneqq 0.06\cos{\frac{7\pi s}{S}}$ where $s = $ step number, $S=$ total number of steps \\
Initial $d_{\textrm{max}}$ & 0.01 \\ \bottomrule
\end{tabular}
\caption{CCP hyperparameter choices.}
\label{tab:ccp_hyperparameters}
\end{table}

\cref{tab:ccp_hyperparameters} details some of the CCP-specific and common hyperparameter choices used in all experiments. 

\subsection{Image Transformations}
\label{subsec:image_transformations}

Each image transformation used in this work features one or more parameters that control the magnitude of its effect. Along with the transformation type, these parameters are randomly varied during training within predefined bounds. All transformations are applied sequentially per image with a probability of occurring. Random crop, horizontal flip, color jitter, and grayscale transformation occurs with probability $100\%$, $50\%$, $80\%$, $20\%$, respectively. A description of each $t\in \mathcal{T}$ can be found below.

\begin{itemize}
    \item \textbf{Random crop}: Take a random crop of the image containing, at a minimum, $10\%$ of the original image. Given that the aspect ratio (width over height) of CIFAR-10 images is $1.0$, The aspect ratio of the crop must fall within $[0.75, 1.25]$. The image crops are resized back to $32 \times 32$ using bicubic interpolation.
    \item \textbf{Horizontal flip}: Flip images left-to-right.
    \item \textbf{Color Jitter}: Randomly distort the brightness, contrast, saturation, and hue of an image in a randomly chosen order. The maximum delta for brightness, contrast, and saturation jitter is set to $0.72$ and $0.18$ for hue jitter.
    \item \textbf{Grayscale}: Transform the colors of the entire image to grayscale.
\end{itemize}

\section{Normalized Data Variable Sensitivity Analysis}
\label{sec:normalized_data_variable_sensitivity_analysis}

\begin{figure}[tb]
  \centering
  \includegraphics[width=1.0\columnwidth]{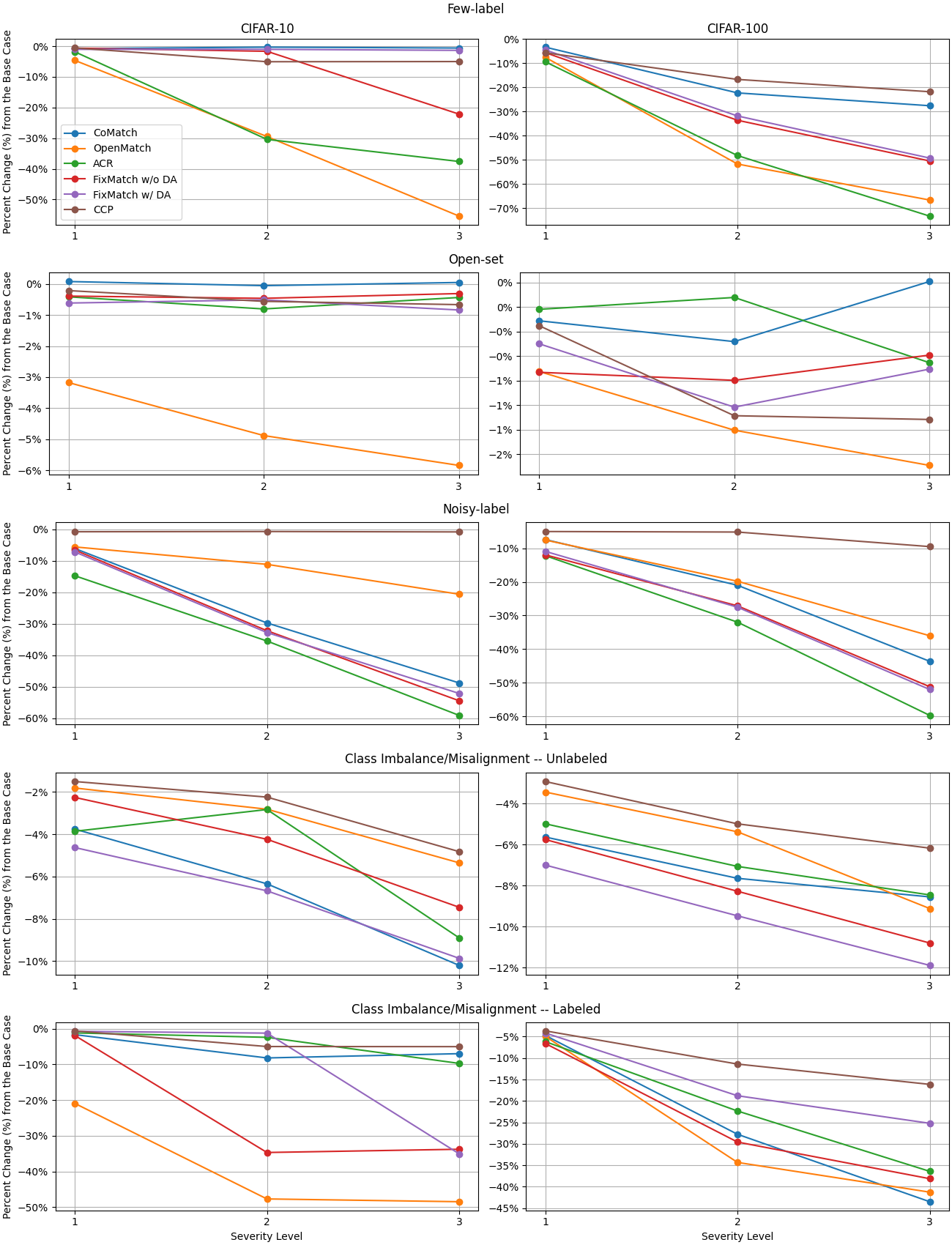}
  \caption{Normalizing the performance of each algorithm in our data variable sensitivity analysis with respect to the corresponding performance on the base case.}
  \label{normalized_plots}
\end{figure}

In \cref{normalized_plots}, we report plots of the performance of each algorithm in our data variable sensitivity analysis that has been normalized with respect to the performance on the base case. This helps control for the effect of inherent label efficiency on the base case such that each data variable can be better analyzed independently.  

\section{Additional Text Classification Experiments}
\label{sec:additional_text_classification_experiments}

As explained in \cref{sec:intro}, our CIFAR experiments focus on testing the reliability of CCP to outperform a supervised baseline in five common SSL data scenarios. We use CIFAR-10 and CIFAR-100 as starting points for those experiments. To ensure CCP works well with a completely different data environment and encoder, we additionally used four well-known text classification datasets: AG News \cite{DBLP:journals/corr/ZhangZL15}, DBpedia \cite{DBLP:journals/corr/ZhangZL15,dbpedia}, Rotten Tomatoes \cite{pang-lee-2005-seeing}, and Yahoo! Answers \cite{10.5555/1620163.1620201,DBLP:journals/corr/ZhangZL15} while investigating the few-label scenario. Some summary information can be found about these datasets in \cref{tab:text_datasets}. 

\begin{table*}[tb]
\centering
\resizebox{0.8\columnwidth}{!}{%
\begin{tabular}{ccccc}
\hline
\textbf{Dataset} & \textbf{No. Classes} & \textbf{Balanced} & \textbf{No. Train Samples} & \textbf{\begin{tabular}[c]{@{}c@{}}Classification\\ Task\end{tabular}} \\ \hline
AG News         & 4  & Yes & 120,000 & Topic                  \\
DBpedia         & 14 & Yes & 560,000 & Topic                  \\
Rotten Tomatoes & 2  & No  & 271,772 & Sentiment              \\
Yahoo! Answers  & 10 & Yes & 700,000 & Topic                  \\ \hline
\end{tabular}%
}
\caption{Description of each text benchmark dataset.}
\label{tab:text_datasets}
\end{table*}

For text data, we compare CCP to a supervised Xent control with access to all labels, a supervised Xent baseline, and three algorithms that bear important similarities to CCP at varying levels of label availability in \cref{tab:text_classification_results}. SimCLR \cite{pmlr-v119-chen20j} and SupCon \cite{https://doi.org/10.48550/arxiv.2004.11362} are included as they are specific cases of $\mathcal{L}_{\text{SSC}}$. Further, the Compact Clustering via Label Propagation (CCLP) regularizer \cite{pmlr-v80-kamnitsas18a} can also be seen as a semi-supervised counterpart of SupCon. Since SimCLR and SupCon require a training session with clean labels only, they should be vulnerable to the few-label scenario. SimCLR and SupCon use the same $f_z$ and $f_g$ that we use with CCP. For CCLP, we attach $f_g$ to $f_b$ and apply CCLP to the hidden layer of $f_g$ for a fair comparison. CCLP calls for linear decision boundaries formed after the hidden space where CCLP is applied via a traditional classification loss applied immediately after a final fully connected layer. We also search over the CCLP-specific hyperparameters such as the number of steps and CCLP weight and report the best results. All networks are trained until convergence according to their originally prescribed procedure. Each CCP experiment consists of 8 CCP iterations with a fixed $\Xi=40$. We use the same subsampling procedure as our CIFAR experiments (initial $d_{\textrm{max}}=0.01$, divided by $2$ after each iteration) between CCP iterations. 

\begin{figure}[tb]
  \centering
  \includegraphics[width=1.0\linewidth]{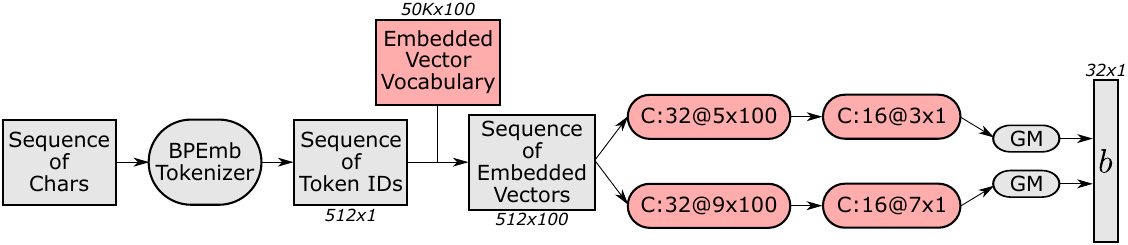}
  \caption{An illustration of an example encoder used in these text classification experiments, $f_b$. ``C:32@5$\times$100" refers to a convolutional layer with 32 filters each of size $5\times 100$. ``GM" refers to global max pooling over the feature activation maps. Red indicates learnable variables.}
  \label{f_b_overview}
\end{figure}

We use a common encoder, $f_b$ for each of the text datasets for all algorithms. The solution is similar to other work that uses convolutional networks for text classification \cite{DBLP:journals/corr/ZhangZL15,kim2014convolutional,NIPS2015_5849}. Briefly, text data is tokenized and each token index is used to look up a corresponding embedded vector (EV). The sequence of EVs is then processed by a convolutional neural network. Unlike our CIFAR experiments, we found no substantial benefit to using self-supervised pretraining to initialize $f_b$ and $f_z$. We use the BPEmb byte-pair encoder \cite{heinzerling2018bpemb} to transform a text document into a sequence of token indices cropped to a certain size. Each token index is used to look up an embedding vector (EV) whose initial values are set to the pretrained BPEmb values. The sequence of EVs is read by parallel convolutional layers whose output then feeds into additional convolutional layers depth-wise. The final activation maps undergo global max pooling to obtain a single floating point value per filter. These maximum activations are concatenated together to form $b_i$. $f_z$ and $f_g$ are designed as a 2-layer MLP. For $f_z$, the size of the hidden (output) layer is 64 (32). For $f_g$ the hidden layer is of size 64 and the output layer size is the number of classes. Small adjustments are made to $f_b$ to ensure the control and baseline reached the desired performance on each text dataset. An illustration of $f_b$ used for text datasets can be found in \cref{f_b_overview}. More information about the hyperparameters of $f_b$ for each dataset can be found in \cref{tab:text_hyperparams}. 

\begin{table*}[tb]
\centering
\resizebox{\columnwidth}{!}{%
\begin{tabular}{@{}cccccc@{}}
\toprule
\textbf{Hyperparameter} &
  \textbf{AG News} &
  \textbf{DBpedia} &
  \textbf{\begin{tabular}[c]{@{}c@{}}Rotten\\ Tomatoes\end{tabular}} &
  \textbf{\begin{tabular}[c]{@{}c@{}}Yahoo!\\ Answers\end{tabular}} \\ \midrule
Input Size         & 512     & 512     & 256     & 1024    \\
Droput Rate        & 0.0     & 0.0     & 0.8     & 0.8     \\
Batch Size         & 256     & 256     & 256     & 256     \\
Optimizer          & Adam    & Adam    & Adam    & Adam    \\
Activation         & GELU    & GELU    & GELU    & GELU    \\
$\tau$             & 0.1     & 0.1     & 0.1     & 0.1     \\
$\Xi$           & 40      & 40      & 40      & 40      \\
Weight Decay       & 0.0     & 0.0     & 0.001   & 0.001   \\
Learning Rate      & 0.00004 & 0.00004 & 0.00004 & 0.00004 \\
\begin{tabular}[c]{@{}c@{}}First Sequence of\\ Parallel Convolutional Layers\end{tabular} &
  32@5$\times$100 $\rightarrow$ 16@3$\times$1 &
  32@5$\times$100 $\rightarrow$ 16@3$\times$1 &
  32@5$\times$100 $\rightarrow$ 16@3$\times$1 &
  64@5$\times$100 $\rightarrow$ 32@3$\times$1 \\
\begin{tabular}[c]{@{}c@{}}Second Sequence of\\ Parallel Convolutional Layers\end{tabular} &
  32@9$\times$100 $\rightarrow$ 16@7$\times$1 &
  32@9$\times$100 $\rightarrow$ 16@7$\times$1 &
  32@9$\times$100 $\rightarrow$ 16@7$\times$1 &
  64@9$\times$100 $\rightarrow$ 32@7$\times$1 \\
\begin{tabular}[c]{@{}c@{}}Third Sequence of\\ Parallel Convolutional Layers\end{tabular} &
  - &
  - &
  - &
  64@13$\times$100 $\rightarrow$ 32@9$\times$1 \\
Global Max Pooling & True    & True    & True    & True    \\ \bottomrule
\end{tabular}%
}
\caption{Hyperparameter settings of $f_b$ for each text dataset. ``32@5$\times$100" refers to a convolutional layer with 32 filters each of size $5\times 100$ \cite{Kingma2015AdamAM, Hendrycks2016}.}
\label{tab:text_hyperparams}
\end{table*}

\subsection{Text Transformations}
\label{subsec:text_transformations}

We design a set of text transformations, $\mathcal{T}$, inspired by the simplest and computationally cheapest set of transformations found in \cite{https://doi.org/10.48550/arxiv.2203.12000}. All transformations are implemented as tensor operations in the computational graph. All transformations act on a sample after it has been converted to a sequence of EVs and allow a gradient to pass through them. These consist of applying Laplacian noise consistent with Differential Privacy \cite{6108143}, applying Gaussian noise, randomly hiding and scrambling the order of EVs, and randomly swapping paragraphs and EVs. The use of transformations alone garners large accuracy increases for every algorithm particularly when label information is scarce. To eliminate this effect on our text experiments, every algorithm tested use the same $\mathcal{T}$ (and augmented batches). Below are more details on each $\mathcal{T}$ used on text:

\begin{itemize}
    \item \textbf{Differential Privacy}: Laplacian noise is applied to all EVs in a fashion that is common in the practice of Differential Privacy \cite{6108143}. We randomly vary the strength of the noise, $\epsilon$, between $10$ and $100$.
    \item \textbf{Gaussian Noise}: Gaussian noise is applied to all EVs. We randomly vary $\mu$ and $sigma$ between $[-0.5, 0.5]$ and $[0.01, 0.05]$, respectively.
    \item \textbf{Vector Hide}: We randomly replace EVs with the learned padding vector used to pad short inputs. The amount of EVs replaced is randomly varied from $10\%$ to $25\%$ of the total length.
    \item \textbf{Paragraph Swap}: We choose a random index along the length of a sequence of EVs and swap the content above and below that index.
    \item \textbf{Random Vector Swap}: We randomly replace EVs with randomly chosen EVs from the full vocabulary. The amount of EVs replaced is randomly varied from $10\%$ to $25\%$ of the total length.
    \item \textbf{Scramble}: We randomly select indices across the length of a sequence of EVs and randomly scramble their order. The amount of EVs we choose to scramble randomly varies from $10\%$ to $25\%$ of the total length.
\end{itemize}

\subsection{Few-Label Performance}
\label{subsec:few_label_performance}

\begin{table}[tb]
\centering
\resizebox{0.7\columnwidth}{!}{%
\begin{tabular}{@{}ccccccc@{}}
\toprule
\textbf{\begin{tabular}[c]{@{}c@{}}\% of\\ Labels\end{tabular}} &
  \textbf{Algorithm} &
  \textbf{DLP} &
  \textbf{AG News} &
  \textbf{DBpedia} &
  \textbf{\begin{tabular}[c]{@{}c@{}}Rotten\\ Tomatoes\end{tabular}} &
  \textbf{\begin{tabular}[c]{@{}c@{}}Yahoo!\\ Answers\end{tabular}} \\ \midrule
100\%                   & Control    & 99.95\%          & 91.95\%          & 98.79\%          & 82.01\%          & 73.17\%          \\ \midrule
\multirow{5}{*}{1\%}    & Baseline   & 99.23\%          & 85.51\%          & 94.25\%          & 66.39\%          & 63.24\%          \\
                        & CCLP       & 99.03\%          & 79.54\%          & 97.57\%          & 64.48\%          & 51.88\%          \\
                        & SimCLR     & 99.46\%          & 86.89\%          & 95.91\%          & 68.10\%          & 57.72\%          \\
                        & SupCon     & 99.40\%          & 85.92\%          & 96.12\%          & 66.97\%          & 60.19\%          \\
                        & CCP (ours) & \textbf{99.70\%} & \textbf{88.89\%} & \textbf{98.24\%} & \textbf{71.14\%} & \textbf{68.26\%} \\ \midrule
\multirow{5}{*}{0.1\%}  & Baseline   & 97.17\%          & 78.37\%          & 86.47\%          & 59.89\%          & 53.44\%          \\
                        & CCLP       & 95.56\%          & 50.66\%          & 67.24\%          & 57.02\%          & 37.80\%          \\
                        & SimCLR     & 97.50\%          & 80.57\%          & 87.75\%          & 59.42\%          & 37.69\%          \\
                        & SupCon     & 97.56\%          & 74.84\%          & 84.97\%          & 57.96\%          & 36.38\%          \\
                        & CCP (ours) & \textbf{98.88\%} & \textbf{87.04\%} & \textbf{96.20\%} & \textbf{60.85\%} & \textbf{63.23\%} \\ \midrule
\multirow{5}{*}{0.05\%} & Baseline   & 94.71\%          & 71.91\%          & 72.71\%          & 57.15\%          & 48.86\%          \\
                        & CCLP       & 91.71\%          & 50.17\%          & 60.13\%          & 55.53\%          & 33.62\%          \\
                        & SimCLR     & 95.32\%          & 77.09\%          & 83.08\%          & 56.07\%          & 38.79\%          \\
                        & SupCon     & 94.96\%          & 68.46\%          & 80.70\%          & \textbf{57.42\%} & 28.25\%          \\
                        & CCP (ours) & \textbf{96.03\%} & \textbf{85.14\%} & \textbf{93.84\%} & 57.35\%          & \textbf{59.58\%} \\ \bottomrule
\end{tabular}%
}
\caption{The best test set accuracy across all text datasets. Bold indicates the best performance found across all algorithms.}
\label{tab:text_classification_results}
\end{table}

Mirroring our CIFAR experiments, note that only CCP outperforms or matches the baseline in every experiment in \cref{tab:text_classification_results}. SimCLR often outperforms the baseline, but CCP outperforms SimCLR in every experiment by the widest margin when label availability is lowest. CCLP underperforms the baseline in every experiment except on DBpedia with $1\%$ label availability \textit{i.e.} when the baseline is highly accurate and the classes are balanced. When CCLP succeeds, it outperforms the pretraining methods likely because it makes direct use of pseudo-labels. CCLP and CCP, like many other SSL algorithms, rely on network fitness for quality pseudo-labels. Accordingly, the margin of success for CCP shrinks as the baseline becomes less accurate \textit{e.g.} on Rotten Tomatoes with $0.05\%$ label availability where the baseline is $<8\%$ better than random guessing.

\subsection{CCP Iteration Performance}
\label{sec:ccp_iter_perf}

In \cref{tab:ccp_iter_perf_full} we display the accuracy of pseudo-labels produced by CCP at each iteration with and without subsampling at all levels of label availability for every text dataset. We can see most runs have converged after seven iterations although minor improvements are still occurring in some runs. Using subsampling can occasionally decrease performance across iterations \textit{e.g.} on the Rotten Tomatoes dataset at $0.1\%$ label availability. After seven iterations, the accuracy is higher than when not using subsampling, but further iterations appear to be decreasing accuracy. This necessitates the careful tuning of a subsampling schedule. Also, occasionally, runs without subsampling appear to return a higher accuracy at termination by a small margin despite a slower start. Again, this is explainable by an accumulation of errors brought forth by subsampling. A smaller initial $d_{\textrm{max}}$ would alleviate this situation. We use a fixed $\Xi=40$ in each CCP iteration for every text dataset. This means all eight CCP iterations consisted of only $320$ epochs.

\begin{table*}[ht]
\centering
\resizebox{\textwidth}{!}{%
\begin{tabular}{@{}cccccccccc@{}}
\toprule
\multirow{2}{*}{\textbf{Dataset}} &
  \multirow{2}{*}{\textbf{\begin{tabular}[c]{@{}c@{}}\% of\\ Labels\end{tabular}}} &
  \multirow{2}{*}{\textbf{Subsampling}} &
  \multicolumn{7}{c}{\textbf{Iteration}} \\ \cmidrule(l){4-10} 
                                 &                         &   & \textbf{1} & \textbf{2} & \textbf{3} & \textbf{4} & \textbf{5} & \textbf{6} & \textbf{7} \\ \midrule
\multirow{6}{*}{AG News}         & \multirow{2}{*}{1\%}    & N & 83.51\%    & 86.68\%    & 87.30\%    & 87.60\%    & 87.79\%    & 87.95\%    & 87.89\%    \\
                                 &                         & Y & 83.51\%    & 86.50\%    & 86.54\%    & 87.23\%    & 87.67\%    & 88.00\%    & 88.13\%    \\ \cmidrule(l){2-10} 
                                 & \multirow{2}{*}{0.1\%}  & N & 75.29\%    & 81.62\%    & 83.27\%    & 84.03\%    & 84.48\%    & 84.67\%    & 84.84\%    \\
                                 &                         & Y & 75.29\%    & 82.17\%    & 83.37\%    & 84.18\%    & 84.74\%    & 85.15\%    & 85.49\%    \\ \cmidrule(l){2-10} 
                                 & \multirow{2}{*}{0.05\%} & N & 67.14\%    & 76.71\%    & 79.66\%    & 80.78\%    & 81.70\%    & 82.27\%    & 82.72\%    \\
                                 &                         & Y & 67.14\%    & 77.56\%    & 80.36\%    & 81.56\%    & 82.00\%    & 82.63\%    & 82.98\%    \\ \midrule
\multirow{6}{*}{DBpedia}         & \multirow{2}{*}{1\%}    & N & 96.24\%    & 97.23\%    & 97.55\%    & 97.69\%    & 97.71\%    & 97.78\%    & 97.81\%    \\
                                 &                         & Y & 96.24\%    & 96.80\%    & 97.06\%    & 97.44\%    & 97.58\%    & 97.66\%    & 97.85\%    \\ \cmidrule(l){2-10} 
                                 & \multirow{2}{*}{0.1\%}  & N & 89.88\%    & 92.84\%    & 93.79\%    & 94.23\%    & 94.58\%    & 94.81\%    & 94.87\%    \\
                                 &                         & Y & 89.88\%    & 93.45\%    & 94.11\%    & 94.42\%    & 94.84\%    & 95.01\%    & 95.25\%    \\ \cmidrule(l){2-10} 
                                 & \multirow{2}{*}{0.05\%} & N & 82.75\%    & 87.43\%    & 89.08\%    & 89.99\%    & 90.57\%    & 90.86\%    & 91.09\%    \\
                                 &                         & Y & 82.75\%    & 88.17\%    & 90.04\%    & 91.06\%    & 91.50\%    & 91.96\%    & 92.26\%    \\ \midrule
\multirow{6}{*}{Rotten Tomatoes} & \multirow{2}{*}{1\%}    & N & 64.29\%    & 67.84\%    & 68.89\%    & 69.42\%    & 69.55\%    & 69.57\%    & 69.70\%    \\
                                 &                         & Y & 64.29\%    & 68.94\%    & 69.92\%    & 70.46\%    & 70.60\%    & 70.66\%    & 70.61\%    \\ \cmidrule(l){2-10} 
                                 & \multirow{2}{*}{0.1\%}  & N & 57.04\%    & 57.75\%    & 58.17\%    & 58.32\%    & 58.38\%    & 58.44\%    & 58.50\%    \\
                                 &                         & Y & 57.04\%    & 58.30\%    & 59.11\%    & 59.34\%    & 59.28\%    & 59.22\%    & 59.14\%    \\ \cmidrule(l){2-10} 
                                 & \multirow{2}{*}{0.05\%} & N & 54.75\%    & 55.28\%    & 55.65\%    & 55.79\%    & 55.89\%    & 55.93\%    & 55.96\%    \\
                                 &                         & Y & 54.75\%    & 55.53\%    & 56.31\%    & 56.41\%    & 56.38\%    & 56.42\%    & 56.41\%    \\ \midrule
\multirow{6}{*}{Yahoo! Answers}  & \multirow{2}{*}{1\%}    & N & 63.06\%    & 65.70\%    & 66.46\%    & 66.88\%    & 66.90\%    & 67.13\%    & 67.22\%    \\
                                 &                         & Y & 63.06\%    & 66.20\%    & 66.43\%    & 66.87\%    & 67.06\%    & 67.30\%    & 67.31\%    \\ \cmidrule(l){2-10} 
                                 & \multirow{2}{*}{0.1\%}  & N & 52.82\%    & 58.23\%    & 59.76\%    & 60.53\%    & 60.99\%    & 61.37\%    & 61.61\%    \\
                                 &                         & Y & 52.82\%    & 58.57\%    & 59.83\%    & 60.32\%    & 60.60\%    & 60.76\%    & 60.92\%    \\ \cmidrule(l){2-10} 
                                 & \multirow{2}{*}{0.05\%} & N & 45.25\%    & 52.94\%    & 55.28\%    & 56.45\%    & 57.19\%    & 57.70\%    & 57.99\%    \\
                                 &                         & Y & 45.25\%    & 53.50\%    & 55.89\%    & 56.98\%    & 57.56\%    & 57.89\%    & 58.13\%    \\ \bottomrule
\end{tabular}%
}
\caption{The accuracy of pseudo-labels after seven iterations of CCP. All runs with subsampling uses an initial $d_{\text{max}}=0.01$.}
\label{tab:ccp_iter_perf_full}
\end{table*}

\section{Pseudo-Label Oscillation}
\label{sec:pseudo_label_oscillation}

\begin{figure}[tb]
  \centering
  \includegraphics[width=1.0\textwidth]{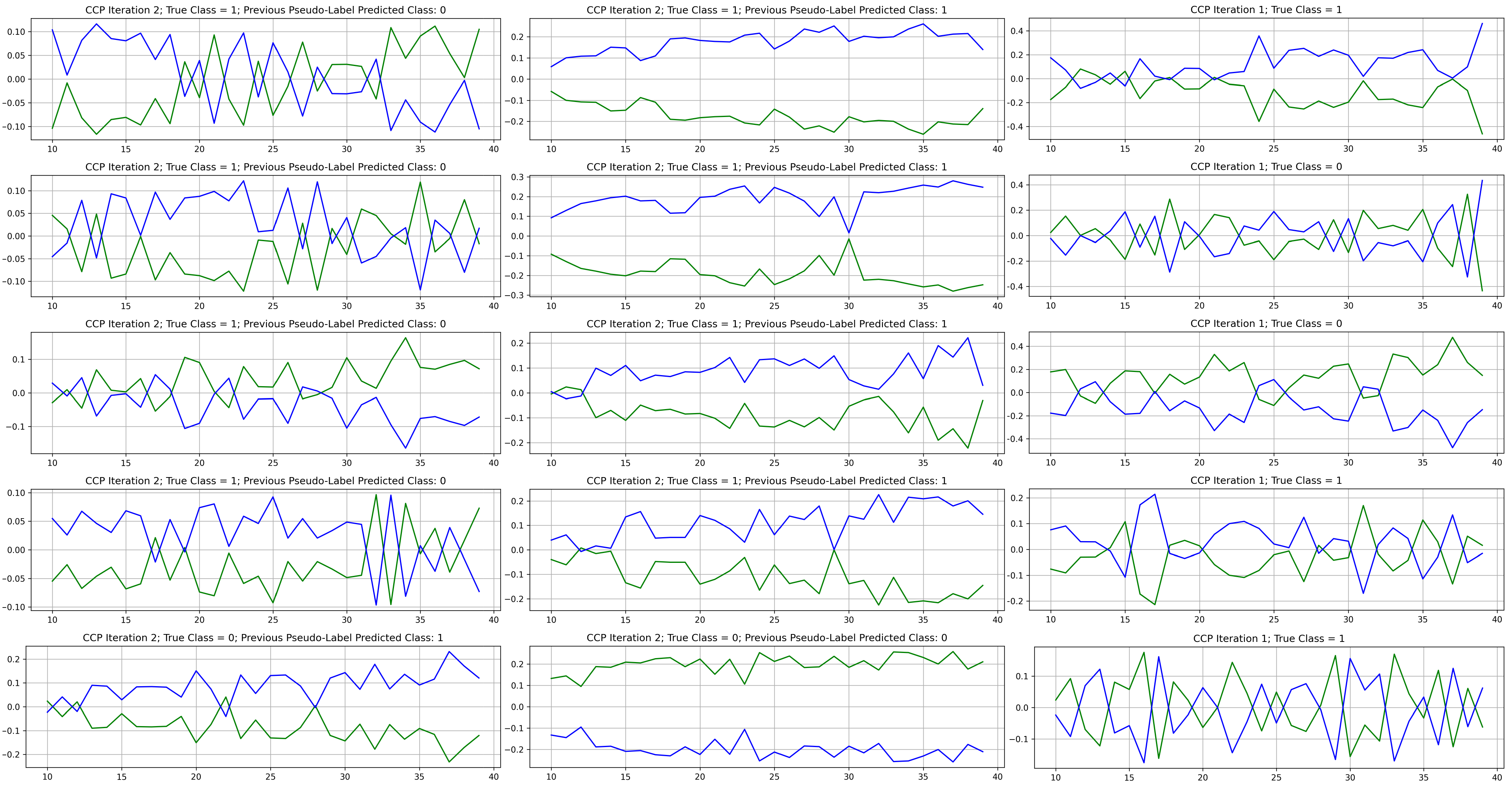}
  \caption{Randomly chosen plots from training CCP with a binary classification dataset (Rotten Tomatoes \cite{pang-lee-2005-seeing}) with $1\%$ label availability depicting pseudo-label oscillation over epochs. On the x-axis is the epoch number. On the y-axis is the non-scaled credibility value for class 1 in blue and class 0 in green. The leftmost column shows oscillation in the presence of a pseudo-label error. The middle column displays significantly degraded oscillation in the presence of a correct pseudo-label. The rightmost column depicts oscillations which can also be observed in the first CCP iteration.}
  \label{oscillation_fig}
\end{figure}

Similar to \cite{https://doi.org/10.48550/arxiv.2012.05458}, we observe oscillations of the predicted pseudo-label across epochs in the presence of a pseudo-label error. The credibility vector value for the correct class begins high (or at least reflects uncertainty \textit{i.e.} near zero) but the network eventually memorizes the incorrect pseudo-label as training continues. When there is no pseudo-label error, the magnitude of oscillation is significantly less. Oscillations are also observed during the first iteration when no prior pseudo-label exists. This is attributable to the effects of random batch selection and network state. Randomly chosen examples of these behaviors are illustrated in \cref{oscillation_fig}. This is empirical evidence for our theoretical justification of CCP iterations in \cref{subsec:the_ccp_iteration}.

\end{document}